\documentclass{article}

\PassOptionsToPackage{numbers, compress}{natbib}
\usepackage[preprint]{neurips_2025}

\usepackage[utf8]{inputenc} 
\usepackage[T1]{fontenc}    
\usepackage{hyperref}       
\usepackage{url}            
\usepackage{booktabs}       
\usepackage{amsfonts}       
\usepackage{afterpage}
\usepackage{nicefrac}       
\usepackage{microtype}      
\usepackage{comment}        
\usepackage{amsmath}
\usepackage{multirow}
\usepackage{bm}
\usepackage{amsthm}
\usepackage{hhline}
\usepackage{amssymb}
\usepackage{makecell}
\usepackage{mathtools}
\usepackage{color}
\usepackage{xcolor}
\usepackage{MnSymbol}
\usepackage{graphicx}
\usepackage{arydshln}
\usepackage{booktabs}
\usepackage{framed}
\usepackage[font=small]{caption}
\usepackage[scientific-notation=true]{siunitx}
\usepackage{wrapfig}
\usepackage{lipsum}
\usepackage{sidecap}
\usepackage{pifont}
\usepackage[flushleft]{threeparttable}
\usepackage{diagbox}
\usepackage{enumitem}
\usepackage{tabularx}
\usepackage{nicefrac}       
\usepackage[parfill]{parskip}
\usepackage{bbm}
\usepackage{cases}
\usepackage{subcaption}
\usepackage{algorithm}
\usepackage{longtable}
\usepackage{algorithmic}
\usepackage{hhline}
\usepackage{xspace}
\usepackage{enumitem}
\usepackage{tikz}
\usepackage{todonotes}
\usepackage{graphicx}

\newtheorem{definition}{Definition}

\newcommand{\model}{\textsc{MINT}}

\newcommand{\fuyu}{\textsc{Fuyu-8B}\xspace}

\newcommand{\kosmostwo}{\textsc{Kosmos-2}\xspace}
\newcommand{\bliptwo}{\textsc{BLIP-2}\xspace}

\newcommand{\openflamingo}{\textsc{OpenFlamingo}\xspace}

\newcommand{\mplugowl}{\textsc{mPLUG-Owl}\xspace}
\newcommand{\VQA}{\textsc{VQA}\xspace}

\newcommand{\FlickrThirtyK}{\textsc{Flickr30K}\xspace}

\newcommand{\NLVR}{\textsc{Nlvr}\xspace}

\newcommand{\IRFL}{\textsc{IRFL}\xspace}
\newcommand{\MMIMDb}{\textsc{MM-IMDb}\xspace}
\newcommand{\MagicBrush}{\textsc{Magic Brush}\xspace}
\newcommand{\NewYorkerCartoon}{\textsc{NY Cartoon}\xspace}
\newcommand{\HatefulMemes}{\textsc{Hateful Memes}\xspace}
\newcommand{\MemeCap}{\textsc{MemeCap}\xspace}
\newcommand{\Memotion}{\textsc{Memotion}\xspace}
\newcommand{\FER}{\textsc{FER-2013}\xspace}
\newcommand{\ScienceQA}{\textsc{ScienceQA}\xspace}
\newcommand{\Resisc}{\textsc{Resisc45}\xspace}
\newcommand{\UCMercedLandUse}{\textsc{UCMerced land use}\xspace}
\newcommand{\iNaturalist}{\textsc{iNaturalist}\xspace}
\newcommand{\Decimer}{\textsc{Decimer}\xspace}
\newcommand{\PathVQA}{\textsc{PathVQA}\xspace}
\newcommand{\VQARAD}{\textsc{VQARAD}\xspace}

\newcommand{\Slake}{\textsc{Slake}\xspace}
\newcommand{\Enrico}{\textsc{Enrico}\xspace}

\newcommand{\LNCOCO}{\textsc{LNCOCO}\xspace}

\title{MINT: Multimodal Instruction Tuning with\\Multimodal Interaction Grouping}

\author{%
  Xiaojun Shan$^{1,*}$, Qi Cao$^{1,*}$, Xing Han$^{2,*}$, Haofei Yu$^{3,*}$, Paul Pu Liang$^{4}$\\
  $^1$University of California, San Diego 
  $^2$Johns Hopkins University\\
  $^3$University of Illinois Urbana-Champaign 
  $^{4}$Massachusetts Institute of Technology \\
  \{xishan, q9cao\}@ucsd.edu, xhan56@jhu.edu, haofeiy2@illinois.edu, ppliang@mit.edu
\\
\textbf{$^{*}$ indicates equal contribution}
}
\begin{document}

\maketitle

\begin{abstract}
Recent advances in multimodal foundation models have achieved state-of-the-art performance across a range of tasks.
These breakthroughs are largely driven by new pre-training paradigms that leverage large-scale, unlabeled multimodal data, followed by instruction fine-tuning on curated labeled datasets and high-quality prompts. While there is growing interest in scaling instruction fine-tuning to ever-larger datasets in both quantity and scale, our findings reveal that simply increasing the number of instruction-tuning tasks does not consistently yield better performance. Instead, we observe that grouping tasks by the common interactions across modalities, such as discovering redundant shared information, prioritizing modality selection with unique information, or requiring synergistic fusion to discover new information from both modalities, encourages the models to learn transferrable skills within a group while suppressing interference from mismatched tasks. 
To this end, we introduce \textbf{\model}, a simple yet surprisingly effective task-grouping strategy based on the type of multimodal interaction. We demonstrate that the proposed method greatly outperforms existing task grouping baselines for multimodal instruction tuning, striking an effective balance between generalization and specialization.

\end{abstract}

\section{Introduction}
\label{sec:introduction}

Multimodal learning has made significant progress in addressing tasks that involve the integration of heterogeneous data sources, such as text, images, and structured knowledge~\citep{baltruvsaitis2018multimodal, wu2023multimodal, li2023blip, radford2021learning}. While large-scale pre-training is undoubtedly crucial~\citep{brown2020language, raffel2020exploring}, state-of-the-art models often follow this phase with task-specific instruction fine-tuning (FT) to adapt to downstream applications~\citep{wei2021finetuned, sanh2021multitask} and better align with real-world human use cases~\citep{ouyang2022training, rafailov2023direct}. The expansive landscape of instruction tuning tasks and datasets~\citep{longpre2023flan, wang2022super} raises important open questions about how best to order, curate, and group tasks for optimal instruction tuning. While earlier work has largely focused on either single-task tuning~\citep{devlin2019bert, raffel2020exploring} or indiscriminate multi-task aggregation~\citep{wei2021finetuned, sanh2021multitask}, emerging research emphasizes the importance of uncovering the latent structure and relationships among tasks. Such understanding can inform the selection of effective tasks for tuning~\citep{zhou2023lima, liu2023makes}, enable better predictions of scaling behaviors~\citep{chung2024scaling, longpre2023flan}, and guide the design of future instruction tuning protocols~\citep{longpre2023flan, zhou2023lima, liu2023makes}.

Existing task-grouping schemes in multitask learning cluster tasks by simple similarities in inputs \citep{shui2019principled}, labels \citep{fifty2021efficiently, li2024identifying}, or gradients \citep{standley2020tasks, navon2022multi}. Such metrics falter in instruction tuning, where tasks appear as open-ended, often multimodal generation with vast input–output spaces and high-dimensional gradients. Recent instruction-tuning heuristics—instruction-based selection \citep{lee2024instruction}, MoE routing \citep{shen2024multimodal, shen2024mome}, or embedding-based clustering \citep{gou2023mixture} also distinguish tasks at a surface-level, ignoring deeper multimodal interactions.

\begin{figure}[t]
    \centering
    \includegraphics[width=\linewidth]{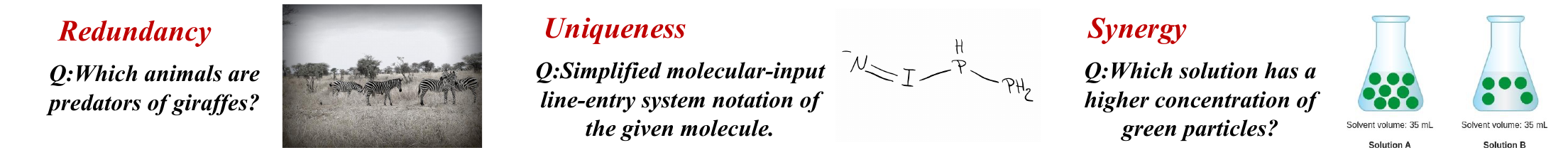}
    \caption{Examples of multimodal instructions exhibiting redundancy, uniqueness, and synergy interactions, as identified by our task grouping method. In the redundancy example, both the text and image indicate a zebra. In the uniqueness example, the text description and the molecular structure provide distinct information. In the synergy example, the text and image jointly formulate a coherent question.}
    \label{fig:rus_demo}
\end{figure}

Rather than relying on similarities in high-dimensional input or label spaces, our key insight is that multimodal tasks often share a latent dimension of skills—the internal capabilities required to solve problems—such as learning to model interactions across modalities~\citep{liang2022foundations, marsh2003taxonomy, kruk2019integrating, bateman2014text}. As shown in Figure \ref{fig:rus_demo}, some tasks require the ability to identify redundant, shared information between modalities~\citep{hwang2023memecap, chhavi2020memotion}; others emphasize selecting the most informative modality when each provides unique content~\citep{he2020pathvqa, Lau_Gayen_Demner_BenAbacha_2019}; and yet others demand synergistic fusion to uncover meaning that emerges only when modalities are combined, such as detecting sarcasm from conflicting textual and visual cues~\citep{cai-etal-2019-multi, liang2023quantifying}.

With this in mind, we introduce \textbf{\model}, a structured task-grouping strategy based explicitly on computational analysis of \textbf{M}ultimodal \textbf{INT}eraction. Clustering tasks by these interaction profiles lets the model exploit coherent learning signals, boosting transfer within a group while suppressing interference from mismatched tasks. To investigate the large-scale efficacy of this approach, we focus on instruction tuning Qwen2-VL~\citep{wang2024qwen2}, a state-of-the-art large multimodal model developed for visual and textual data. 
On the large-scale HEMM benchmark~\citep{liang2024hemm} with more than 30 vision-language tasks, \model\ sets new state-of-the-art results on all of them, with performance gains as high as \textbf{26.7\%} on tasks with redundancy, \textbf{17.1\%} on tasks with synergy, and \textbf{24.9\%} on tasks with unique information.
Interaction-aware fine-tuning surpasses both single-task, naive multi-task baselines and other grouping methods. For example, co-training on synergistic reasoning datasets - MemeCap \citep{hwang2023memecap}, Hateful Memes \citep{kiela2020hateful}, and MM-IMDb \citep{arevalo2017gated} - helps the network distill general multimodal reasoning patterns, yielding consistent gains across that category. We release our code and models at \href{https://anonymous.4open.science/r/3863-38D2/}{https://github.com/sxj1215/MINT} to enable larger-scale studies of multimodal foundation models.

\section{Related Work}
\label{sec:related}

\textbf{Multimodal Foundation Models} 
are emerging as significant for future AI, showcasing strong reasoning~\citep{lu2022learn}, interactive dialogue~\citep{koh2023grounding}, and few-shot generalization capabilities~\citep{tsimpoukelli2021multimodal}. These models are often pre-trained using image-text self-supervised learning and subsequently fine-tuned for specific tasks~\citep{li2019visualbert,lu2019vilbert,su2019vl,liang2022highmmt, liang2025lorasculpt}, or they adapt language models with visual input for image-conditioned text generation~\citep{li2022blip, wang2022git}, with cross-modal transformer architectures being a favored backbone due to their adaptability to both language and image data~\citep{chen2020uniter,tsai2019multimodal, han2024fusemoe, liu2024nvila}. The advancement of multimodal foundation models has been significantly propelled by the creation of comprehensive benchmarks covering numerous modalities and tasks~\citep{lee2023holistic,liang2021multibench,gkoumas2021makes,ferraro-etal-2015-survey, dai2025climb}, including recent benchmarks designed to test their capabilities, such as HEMM~\citep{liang2024hemm}, MMMU~\citep{yue2024mmmu}, MME~\citep{fu2023mme}, MMBench~\citep{liu2023mmbench}, LVLM-ehub~\citep{xu2023lvlm}, SEED-Bench~\citep{li2023seed}, Touchstone~\citep{bai2023touchstone}, Mm-vet~\citep{yu2023mm}, ReForm-Eval~\citep{li2023reform}, VisIT-Bench~\citep{bitton2023visit}, and FLAVA~\citep{kiela2022grounding}, as well as benchmarks focused on assessing hallucination~\citep{cui2023holistic} and applications in specialized fields like medicine~\citep{yan2023multimodal} and autonomous driving~\citep{wen2023road}.
Building upon these advancements, MultiModal Instruction Tuning (MMIT) has emerged as a critical research area for enhancing the ability of these models to follow diverse, open-ended instructions that span multiple modalities. This involves training models on datasets composed of instructions paired with multimodal inputs and desired outputs, thereby improving their zero-shot and few-shot generalization to novel tasks and instructions \citep{liu2023visual}. Given the growing volume of multimodal tasks across diverse research domains, common patterns of interaction often emerge within and across modalities and datasets. Leveraging these intrinsic interactions can significantly enhance the effectiveness of fine-tuning. Consequently, developing robust methods to identify and group tasks for joint instruction tuning is a crucial research direction with the potential to improve both generalization and task-specific performance.

\begin{figure}[t]
    \centering
    \includegraphics[width=\linewidth]{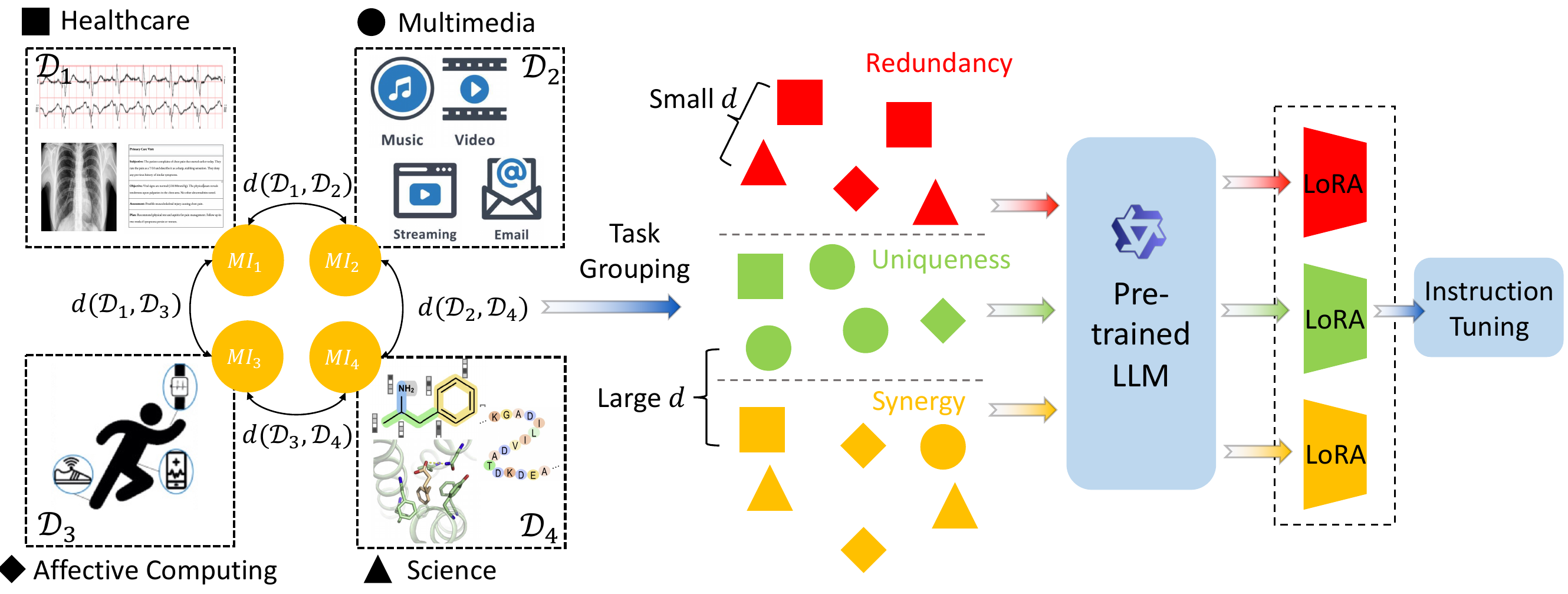}
    \caption{Overview of our approach: For each multimodal instruction tuning dataset, spanning four domains: healthcare (\(\blacksquare\)), multimedia (\ding{108}), affective computing (\(\blacklozenge\)), and science (\(\blacktriangle\)), we first compute its multimodal interaction (MI) score, along with the pairwise multimodal dataset distance (MDD). These scores reflect the dominant interaction type (redundancy, uniqueness, or synergy) and quantify interaction-based dissimilarity between datasets. We then group tasks based on their MI characteristics: datasets within the same group exhibit similar dominant interactions and have low MDD, while datasets across different groups show larger MDD. These interaction-based groupings are subsequently used to perform joint instruction fine-tuning of pre-trained LLMs, promoting better alignment between model capabilities and task demands.}
    \label{fig:overview_fig}
\end{figure}

\paragraph{Task Grouping Methods}
There has been substantial work on grouping tasks for multi-task learning. Foundational work by \citep{caruana1997} showed that learning related tasks in parallel with shared representations improves generalization through inductive transfer. Formalizing this insight, \citep{evgeniou2004regularized} introduced a kernel-based regularization framework in which task parameters are encouraged to cluster around a common mean, although choosing the coupling strength remains challenging. Extending these ideas to contextual bandits, \citep{deshmukh2017multi} leveraged estimated task similarity to obtain tighter regret bounds and better empirical performance than either independent or fully pooled learners. For sequence-labeling tasks, \citep{bingel2017identifying} found that the benefits of multi-task learning with deep neural networks can be predicted from dataset statistics and single-task learning curves. \citep{ma2018modeling} proposed the Multi-gate Mixture-of-Experts architecture, which uses task-specific gating networks over shared experts to learn a dynamic sharing mechanism, albeit without producing explicit task groupings.
Mitigating negative transfer has been another focus \citep{zhang2022transfer, gao2023enhancing}. \citep{yu2020gradient} introduced gradient surgery, a model-agnostic method that projects conflicting task gradients onto each other’s normal planes, markedly improving efficiency and performance. TAG \citep{fifty2021efficiently} determines task groupings by measuring inter-task affinity through gradient interactions during a single joint training run, matching the performance of far more expensive search methods. A systematic comparison by \citep{peters2019tune} showed that the relative effectiveness of feature extraction and fine-tuning depends on the similarity between the pre-training and target tasks, underscoring the central role of task-relatedness. Theoretical support for these empirical findings is provided by \citep{shui2019principled}, who derived generalization bounds that quantify the benefits of incorporating explicit and implicit task similarity information. In multimodal settings, \citep{liang2022highmmt} quantified modality and interaction heterogeneity via information transfer to guide dynamic parameter sharing. For instruction tuning, \citep{lee2024instruction} selected source tasks based on instruction-text similarity, offering an efficient strategy that bypasses instance-level data yet may rely on surface-level cues.

\section{Methodology}

We discuss detailed steps of our method, which (1) clusters tasks by their redundancy, uniqueness, or synergy-dominant multimodal interaction scores, followed by (2) instruction fine-tuning of pretrained models on each cluster. This coherent grouping fosters shared reasoning, avoids cross-task interference, and directs fine-tuning updates toward interaction-specific specialisation.  
\newcommand{\MIscore}{\overline{\Delta}_{1,2}}
\newcommand{\datasetdist}{d}

\subsection{Dataset-Level Multimodal Interactions}\label{sec:data_mi}
Multimodal interaction (MI) is a core capability of multimodal models~\citep{liang2024hemm}. Prior work identifies three prototypical MI types—\textit{redundant}, \textit{unique}, and \textit{synergistic}—drawing from information-theoretic frameworks~\citep{williams2010nonnegative, liang2023quantifying}. 

\citet{yu2024mmoe} further proposes to approximate interaction types via prediction similarity between unimodal and multimodal models. Building on this insight, we generalize the approach by introducing a \textit{dataset-level interaction} score, enabling principled comparisons and groupings of multimodal tasks based on their interaction characteristics.

Denote $\mathcal{X}_1$ and $\mathcal{X}_2$ as feature spaces for two modalities, and $\mathcal{Y}$ the semantic output space. Let $f_1: \mathcal{X}_1 \rightarrow \mathcal{Y}$ and $f_2: \mathcal{X}_2 \rightarrow \mathcal{Y}$ be unimodal models, and $f_m: \mathcal{X}_1 \times \mathcal{X}_2 \rightarrow \mathcal{Y}$ a multimodal model. Define a semantic similarity function $\delta: \mathcal{Y} \times \mathcal{Y} \rightarrow [0, 1]$, where $\delta(y_a, y_b) = 1$ denotes identical semantics and $\delta(y_a, y_b) = 0$ denotes complete dissimilarity.

\vspace{1mm}
\begin{definition}[Multimodal Dataset Distance]
Let $\mathcal{D}$ be a given multimodal dataset. We perform $S$ draws on $\mathcal{D}$: for each draw $s$, we sample a fixed size of $C$ instances, denoted as $\mathcal{D}^{(s)} = \{(x_{1,j}^{(s)}, x_{2,j}^{(s)})\}_{j=1}^{C}$. Here, $(x_{1,j}^{(s)}, x_{2,j}^{(s)})$ are the inputs from the two modalities for that sample. The MI score for $\mathcal{D}$ is calculated as:
\begin{equation}
\overline{\Delta}_{1,2}(\mathcal{D}) = \frac{1}{S} \sum_{s=1}^{S} \overline{\Delta}_{1,2}(\mathcal{D}^{(s)}) = \frac{1}{S} \sum_{s=1}^{S} \left( \frac{1}{C} \sum_{j=1}^{C} \left[ \delta(y_{1,j}^{(s)}, y_{m,j}^{(s)}) + \delta(y_{2,j}^{(s)}, y_{m,j}^{(s)}) \right] \right)
\label{eq:mi_score}
\end{equation}
Given two multimodal datasets $\mathcal{D}_A$ and $\mathcal{D}_B$, their dataset distance is defined as:
\begin{equation}
d(\mathcal{D}_A, \mathcal{D}_B) = \left| \overline{\Delta}_{1,2}(\mathcal{D}_A) - \overline{\Delta}_{1,2}(\mathcal{D}_B) \right|.
\label{eq:data_distance}
\end{equation}
\label{def:def1}
\end{definition}
Here, the MI interaction score $\overline{\Delta}_{1,2}(\mathcal{D})$ quantifies the average agreement between unimodal and multimodal predictions on dataset $\mathcal{D}$; its value provides insights into the interaction type of a dataset. The dataset distance $d(\mathcal{D}_A,\mathcal{D}_B)$ captures the fundamental interaction-based dissimilarity between datasets, enabling principled task grouping and informed multi-task design.

\subsection{Task Grouping in Multimodal Instruction Tuning} \label{sec:task_grouping}
Let $\mathcal{M}$ be a pre-trained multimodal foundation model with parameters $\Theta_0$.
Let $\mathcal{T} = \{\mathcal{T}_1, \mathcal{T}_2, \dots, \mathcal{T}_K\}$ be a set of $K$ distinct multimodal instruction tuning tasks. Each task $\mathcal{T}_k$ is associated with a dataset $\mathcal{D}_k = \{(I_j^k, Q_j^k, A_j^k)\}_{j=1}^{N_k}$, where $I_j^k$ is the multimodal input, $Q_j^k$ is the instruction or question, and $A_j^k$ is the desired output or answer.
The general task grouping problem for instruction tuning is to find a partition $\mathcal{P} = \{G_1, G_2, \dots, G_M\}$ of the set of tasks $\mathcal{T}$, where each $G_m$ is a group of tasks, such that $\bigcup_{m=1}^{M} G_m = \mathcal{T}$ and $G_m \cap G_{m'} = \emptyset$ for $m \neq m'$. 

The goal of this grouping is to improve the overall performance and efficiency of the instruction tuning process. Specifically, by jointly fine-tuning the model $\mathcal{M}$ on tasks within the same group $G_m$, we aim to: (1) \textbf{maximize positive knowledge transfer} among related tasks within a group. (2) \textbf{minimize negative interference} that might occur when unrelated tasks are tuned together, and (3) \textbf{improve generalization} to unseen examples of tasks within the same group.
Let $L(A_j^k, \mathcal{M}(I_j^k, Q_j^k; \Theta))$ be the loss function for a given instance $(I_j^k, Q_j^k, A_j^k)$ when the model $\mathcal{M}$ has parameters $\Theta$.
If we fine-tune the model on a group $G_m$, the objective is to find parameters $\Theta_m^*$ that minimize the aggregate loss for tasks in that group:
\begin{equation}
\Theta_m^* = \arg\min_{\Theta} \sum_{\mathcal{T}_k \in G_m} \sum_{j=1}^{N_k} L(A_j^k, \mathcal{M}(I_j^k, Q_j^k; \Theta)).
\label{eq:objective}
\end{equation}
The challenge lies in defining the criteria for forming the groups $G_m$ such that this objective is best achieved. Indiscriminate multi-task learning can lead to negative interference, while single-task tuning limits generalization. We address this by introducing an explicit task grouping strategy based on the fundamental nature of multimodal interactions required by each task. This approach leverages the Multimodal Dataset Distance derived from interaction characteristics to group tasks.

\subsection{Proposed Method} \label{sec:proposed}

We divide our method into distinct stages: tasks are explicitly grouped based on their dominant interaction type; the model is then fine-tuned separately on these coherent task groups. 

\textbf{Step 1: MI Score and Categorization}~
For each dataset $\mathcal{D}_k$, we first compute its MI score $\MIscore(\mathcal{D}_k)$, as defined in \eqref{eq:mi_score}, which ranges from 0 to 2. Tasks can be categorized based on this score:
\begin{itemize}[leftmargin=8pt,labelsep=0.3em]
\item \textbf{Redundancy-dominant ($G_R$)}: $\MIscore(\mathcal{D}_k) \approx 2$. Information is largely duplicated across modalities.
\item \textbf{Uniqueness-dominant ($G_U$)}: $\MIscore(\mathcal{D}_k) \approx 1$. Critical information is present in only one modality.
\item \textbf{Synergy-dominant ($G_S$)}: $\MIscore(\mathcal{D}_k) \approx 0$. New information emerges from combining modalities.
\end{itemize}

\textbf{Step 2: Explicit Task Grouping}~
We compute the Multimodal Dataset Distance between two datasets $\mathcal{D}_A$ and $\mathcal{D}_B$ as defined in \eqref{eq:data_distance}.
Tasks are grouped such that tasks within the same group have small pairwise dataset distances, meaning they share similar MI scores and thus similar interaction demands. This forms an explicit partition $\mathcal{P}_{RUS} = \{G_R, G_U, G_S\}$ of $\mathcal{T}$.

\textbf{Step 3: Group-Specific Instruction Tuning}~
For each task group $G_m \in \mathcal{P}_{RUS}$:
an instance of the model $\mathcal{M}$ is fine-tuned solely on the aggregated data from tasks within $G_m$. This results in specialized model parameters $\Theta_R^*, \Theta_U^*, \Theta_S^*$.

\paragraph{Advantages of Interaction-Based Grouping} 
Our method groups multimodal tasks by their RUS interaction, steering instruction‑tuning toward the skills each interaction demands. Training jointly on tasks that share an interaction type repeatedly exercises the same mechanisms—cross‑modal fusion for synergy, selective attention for uniqueness, and redundant cross‑checking—so the model distils domain‑agnostic reasoning patterns and transfers them across domains. Separating RUS types also prevents negative interference: updates learned for redundancy no longer erode the fine‑grained fusion required by synergy, because each batch delivers coherent gradients. For large foundation models, whose parameters are only lightly nudged during fine‑tuning, such principled organization is crucial: it channels small updates into purposeful modulation of pre‑trained representations, sharpens cross‑modal alignment strategies, and respects the language‑conditioned cues on which MMIT tasks rely. Unlike similarity measures based on labels or surface statistics, RUS captures the core computational demands of a task, giving the model clear signals about how to process information rather than what content to memorize, and yielding consistently stronger adaptation and generalization.

\subsection{Instruction Tuning Steps on Task Groups} \label{sec:tuning_step}

\textbf{Data Preparation}~
Once the task groups are identified, the next step is to prepare the data for instruction tuning Qwen2-VL. This involves formatting the input for each question within a group, typically including the question, any associated context (text or image), and the multiple-choice options. The desired output during fine-tuning would be the correct answer. We adopt the ShareGPT conversational format, in which each example is represented as a sequence of alternating ``user'' and ``assistant'' messages. For every dataset instance, the ``user'' turn contains:
\begin{itemize}
  \item \textbf{Question prompt} (e.g. “What emotions does this image convey?”),  
  \item \textbf{Associated context}—either free-form text, an image URL or both—and  
  \item \textbf{Answer options} (when applicable, formatted as “A. \dots\  B. \dots\  C. \dots\  D. \dots”).
\end{itemize}

To maintain the instructional integrity and balance of our fine‐tuning corpus, we explicitly exclude two types of datasets: (1) Most VQA–only benchmarks—this ensures that our tuning data emphasizes multimodal instruction understanding across diverse tasks—retrieval, classification, reasoning, and open-ended generation—rather than low-level visual question answering; and (2) some sources likely seen during Qwen‐VL2’s pretraining or whose content overlaps substantially with our selected splits, since such familiar examples contribute little new instructional signal and risk fostering rote memorization rather than genuine multimodal reasoning. By filtering out both purely VQA tasks and redundant corpora, we ensure that every remaining dataset delivers diverse challenges aligned with our target use cases.

\textbf{Fine-Tuning}~
With the data prepared for each task group, we proceed with the instruction tuning process on the Qwen2-VL model. This will involve selecting appropriate hyperparameters such as the learning rate, batch size, and the number of training epochs. Given the size of Qwen2-VL, employing parameter-efficient fine-tuning (PEFT) techniques like Low-Rank Adaptation (LoRA)\citep{hu2022lora} could be beneficial to reduce computational costs and memory requirements while still allowing for effective adaptation to the specific characteristics of each task group. 

\textbf{Evaluation}~
After fine-tuning Qwen2-VL on each task group, the model's performance will be evaluated on a test set of examples from the tasks within that group. Additionally, we will compare the performance against baseline conditions, such as fine-tuning Qwen2-VL on each individual task separately and fine-tuning it on all tasks within the chosen datasets group together, without any specific grouping. This comparative evaluation will allow us to assess whether fine-tuning on groups of similar tasks, as defined by their hypothesized multimodal interaction types, indeed leads to superior performance.

\section{Experiments}
\label{sec:exps}

\begin{table*}[]
\fontsize{9}{10}\selectfont
\setlength\tabcolsep{2.5pt}
\vspace{-4mm}
\caption{We select 18 representative datasets from HEMM \citep{liang2024hemm}, which provide a comprehensive suite to benchmark multimodal foundation models. We categorize each dataset based on the \textit{basic multimodal skills} needed to solve them -- the type of multimodal interaction.}
\centering
\vspace{-0mm}
\begin{tabular}{lccc}
\toprule
Dataset & \# Samples & Interactions & Use case \\
\midrule
\FlickrThirtyK~\citep{flickr30k} & 30K & Redundancy & Multimedia \\
\NLVR~\citep{suhr2017corpus} & 92K & Redundancy & Multimedia \\
\IRFL~\citep{yosef2023irfl} & 3.9K & Synergy & Multimedia \\
\MMIMDb~\citep{arevalo2017gated} & 25K & Synergy & Multimedia \\
\NewYorkerCartoon~\citep{hessel2022androids} & 364 & Synergy & Affect \\
\HatefulMemes~\citep{kiela2020hateful} & 10K & Synergy & Affect \\
\MemeCap~\citep{hwang2023memecap} & 560 & Synergy & Affect \\
\Memotion~\citep{chhavi2020memotion} & 10K & Synergy & Affect \\
\FER~\citep{goodfellow2013challenges} & 30K & Uniqueness & Affect \\
\ScienceQA~\citep{lu2022learn} & 21K & Synergy & Science \\
\Resisc~\citep{cheng2017remote} & 31K & Uniqueness & Science \\
\UCMercedLandUse~\citep{yang2010bag} & 2K & Uniqueness & Science \\
\iNaturalist~\citep{van2017inaturalist} & 675K & Uniqueness & Science \\
\Decimer~\citep{brinkhaus2022decimer} & 5K & Uniqueness & Science \\
\PathVQA~\citep{he2020pathvqa} & 33K & Redundancy & Healthcare \\
\VQARAD~\citep{lau2018dataset} & 3.5K & Redundancy & Healthcare \\
\Slake~\citep{liu2021slake} & 13K & Redundancy & Healthcare \\
\Enrico~\citep{leiva2020enrico} & 1.4K & Uniqueness & HCI \\
\bottomrule
\end{tabular}
\vspace{-4mm}
\label{data:overview}
\end{table*}

\subsection{Experimental Setup}

\subsubsection{Datasets} 
We use the datasets in the HEMM~\citep{liang2024hemm} benchmark. As stated in the previous section, since the distribution of HEMM is imbalanced, i.e. different interaction type has various numbers of tasks, and also we assume that some datasets are commonly used in pertaining, like VQA~\citep{antol2015vqa} and GQA~\citep{hudson2019gqa}, which can cause bias in instruction tuning. Thus, we select part of the HEMM as our training dataset. After computing the Multimodal Dataset Distance, we obtained three distinct groups:
\begin{itemize}[leftmargin=8pt,labelsep=0.3em]
  \item \textbf{Redundancy}: \textsc{slake, pathvqa, vqarad, ok-vqa, nlvr, flickr30k}
  \item \textbf{Synergy}: \textsc{mmimdb, memecap, hateful\_memes, ny\_cartoon, memotion, scienceqa}
  \item \textbf{Uniqueness}: \textsc{enrico, fer2013, resisc45, decimer, ucmerced, inaturalist}
\end{itemize}

For testing, we selected 12 validation sets of the data set seen in the training set, like SLAKE~\citep{liu2021slake} or unseen in the training set, like MagicBrush~\citep{Zhang2023MagicBrush} from HEMM for fair comparison. They are:
\begin{itemize}[leftmargin=8pt,labelsep=0.3em]
  \item \textbf{Redundancy}: \textsc{slake, PathVQA, VQA, nlvr}
  \item \textbf{Synergy}: \textsc{HatefulMemes, NyCartoon, MagicBrush, ScienceQA}
  \item \textbf{Uniqueness}: \textsc{lncoco, inaturalist, Screen2Words, ucmerced}
\end{itemize}

\subsubsection{Model}


The \textbf{Qwen2-VL} model \citep{wang2024qwen2} consists of a 675M-parameter Vision Transformer (ViT) paired with Qwen2 language models of varying scales (2B, 7B, and 72B). To balance effectiveness and efficiency, we performed instruction tuning on the 7B scale Qwen2-VL model. A major innovation is its Naive Dynamic Resolution mechanism, which allows the model to process images of arbitrary resolutions by dynamically adjusting the number of visual tokens. This is enabled by replacing fixed positional embeddings in the ViT with 2D Rotary Position Embeddings, which capture two-dimensional spatial relationships. After tokenization, adjacent $2 \times 2$ image patches are compressed into single tokens using a lightweight MLP, and the visual token sequence is enclosed with \texttt{<|vision\_start|>} and \texttt{<|vision\_end|>} markers before being passed into the language model.

\subsubsection{Baseline}

Our key experiments compare the performance of Qwen2-VL under three main settings:
\begin{itemize}[leftmargin=8pt,labelsep=0.3em]
    \item \textbf{Single-Task Fine-Tuning:} Fine-tuning the model on data from each individual dataset.
    \item \textbf{Unselective Multi-Task Fine-Tuning:} Fine-tuning the model on the combined data from all tasks among all datasets, without any explicit grouping.
    \item \textbf{\model}: Fine-tuning the model separately on the data from each of the task groups defined based on the same multimodal interaction types.
\end{itemize}

To assess the advantage of our grouping mechanism, we compare \model\ against two alternative strategies: (1) INSTA \citep{lee2024insta}: grouping tasks by instruction-text similarity, and (2) MixLora \citep{shen2024multimodal}: dynamically allocating model capacity using conditioned on cluster assignments. 
INSTA leverages a Sentence Transformer to compute cosine similarities between task‐instruction embeddings, allowing it to automatically identify and select the most pertinent source tasks for a given target without requiring raw data samples or exhaustive pairwise transfer experiments. To better align with the meta‐dataset’s instructional style, the similarity model is further fine‐tuned via supervised contrastive learning. Applying INSTA, we partition the datasets into three clusters:

\begin{itemize}[leftmargin=8pt,labelsep=0.3em]
  \item \textbf{Group 1:} \textsc{PathVQA, VQARAD, Hateful Memes, MemeCap, Memotion, NYCartoon}
  \item \textbf{Group 2:} \textsc{RESISC45, UCMerced, FER-2013, ScienceQA, MM-IMDb, Screen2Words}
  \item \textbf{Group 3:} \textsc{Slake, OK-VQA, Enrico, Decimer, Flickr30K, iNaturalist}
\end{itemize}

MixLora introduces the Conditional Mixture‐of‐LoRA framework for multimodal instruction tuning. In MixLoRA, each transformer layer contains a small pool of low‐rank adapters, and a lightweight gating network computes instance‐specific mixture weights over these adapters based on the instruction embedding. This design reduces trainable parameters while enabling adapter specialization.

In our work, we preserve all MixLoRA hyperparameters (adapter pool size $E=16$, rank $r=4$, learning rate, batch size, epochs) but replace the original LLaVA‐v1 backbone with Qwen2-VL. This swap required non‐trivial efforts to align LoRA adapter dimensions, reconfigure the gating‐network inputs and modality‐fusion interfaces, and adapt to Qwen2-VL’s distinct encoder architecture. 

\subsection{Experimental Results}

\begin{figure}[t]
  \centering
  \includegraphics[width=\textwidth]{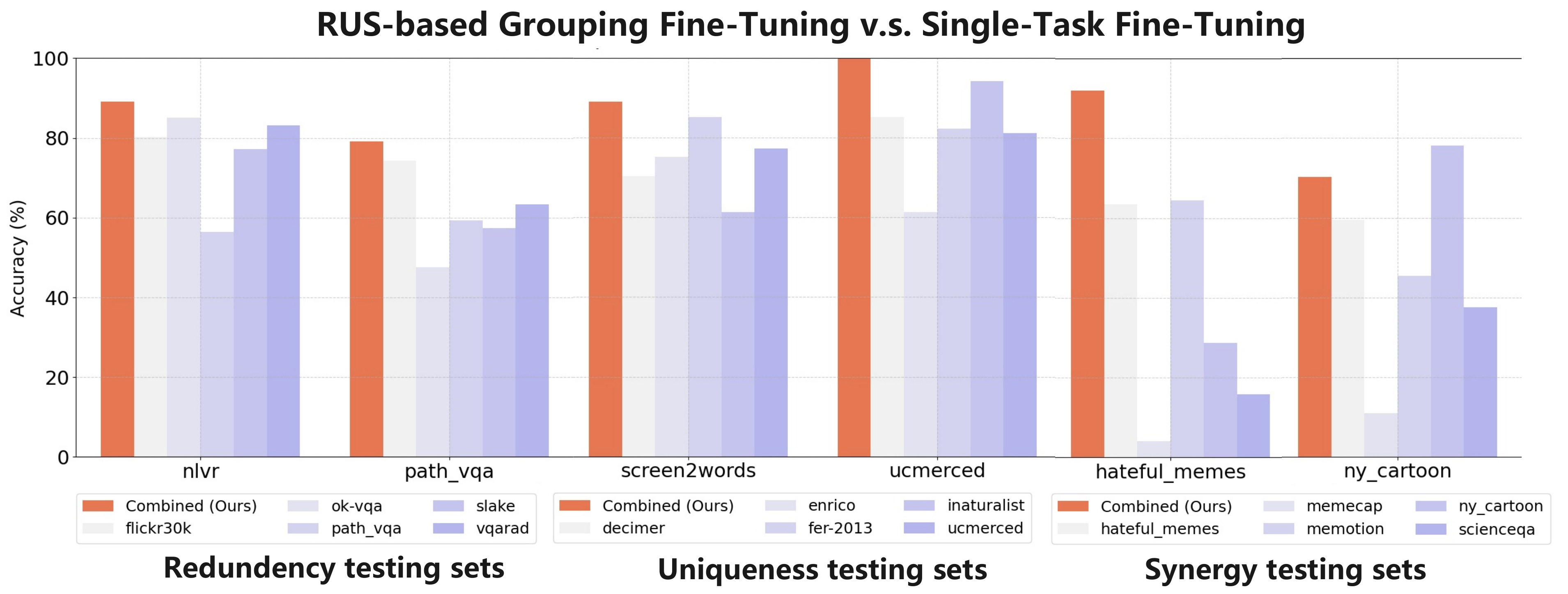} 
  \caption{\textbf{Single-Task Fine-Tuning Comparison.} We compare our \model\ method against models that are fine-tuned individually on each dataset.}
  \label{fig:single}
\end{figure}

\begin{figure}[t]
  \centering
\begin{subfigure}[t]{0.49\textwidth}
    \centering
    \includegraphics[width=\linewidth]{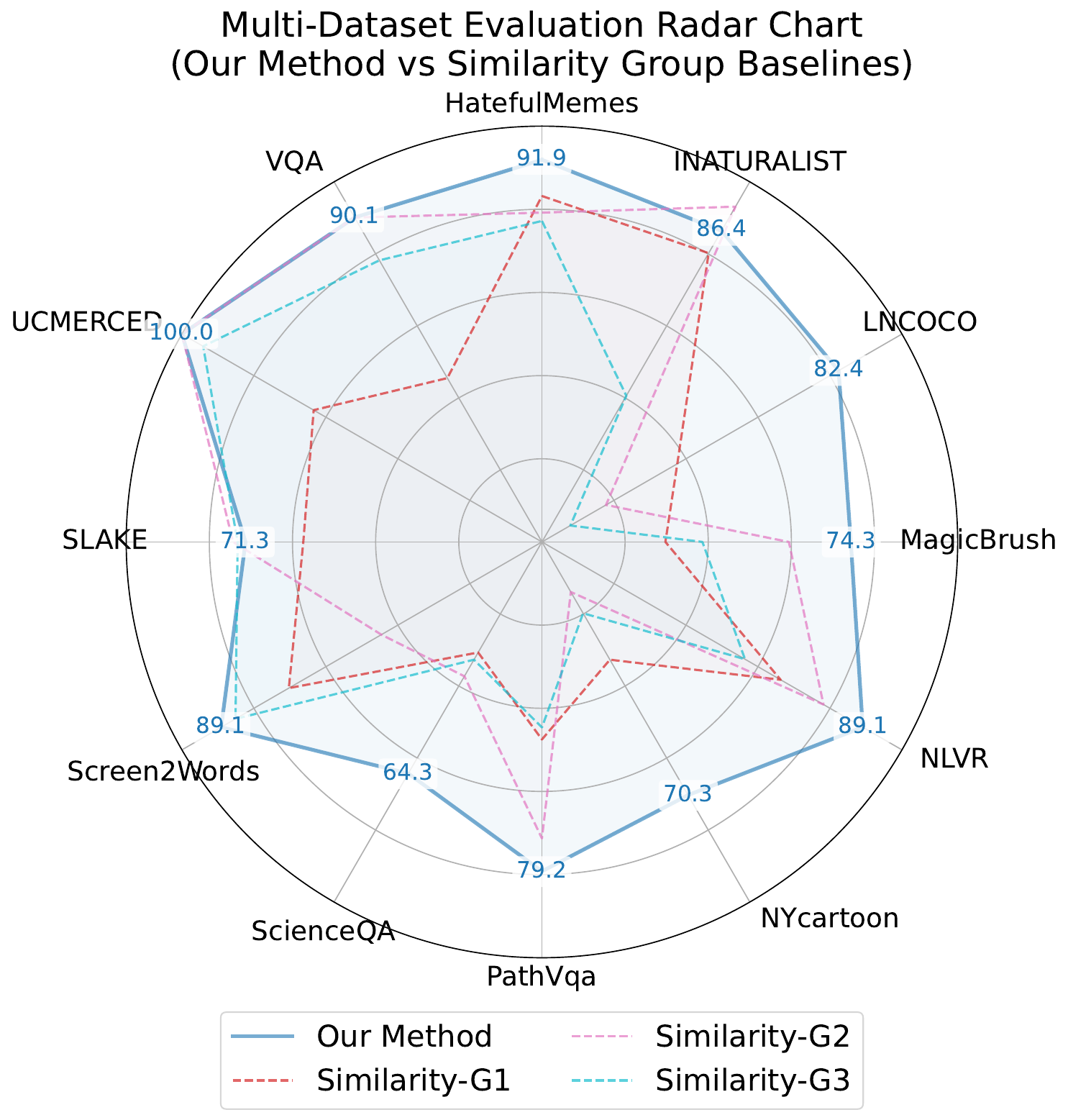}

    \label{fig:radar1}
  \end{subfigure}
  \hfill
  \begin{subfigure}[t]{0.49\textwidth}
    \centering
    \includegraphics[width=\linewidth]{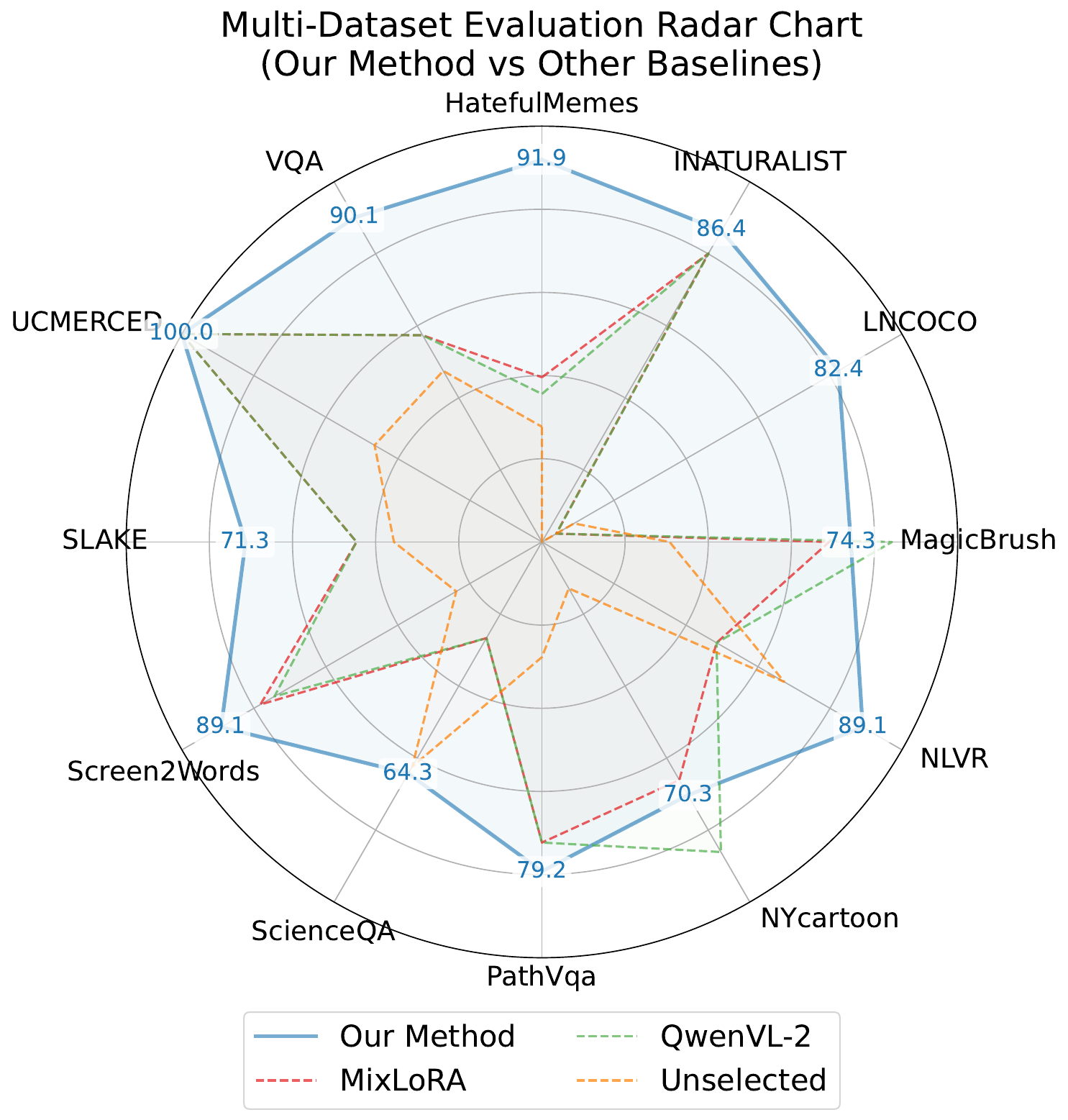}
    \label{fig:radar2}
  \end{subfigure}
  \caption{
\textbf{Cross-Dataset Performance Comparison.}
The radar chart reports accuracies (\%) on twelve benchmarks, where the outer rim represents the best score per dataset.
The solid blue line is \model\.
The dotted pink, cyan, and magenta curves (\textit{Similarity-G1 to Similarity-G3 fine-tuned}) correspond to \emph{Similar-Task Group Fine-Tuning} baselines, each trained on a cluster of related datasets.
The dashed green curves (\textit{Unselective}) are \emph{Unselective Multi-Task Fine-Tuning} baselines, trained on the full dataset mixture. The dashed red and orange curves correspond to MixLoRA fine-tuning baseline and Qwen2-VL Model without being fine-tuned.
Our Method achieves the highest or second-highest accuracy on the majority of datasets and delivers the best average performance, highlighting the benefit of dynamically reweighting domains instead of either coarse group selection (G1–G3) or uniform multi-task training (All).
}
  \label{fig:radar}
\end{figure}

\subsubsection{Effectiveness of \model}
We first evaluate the impact of instruction fine-tuning under different data grouping strategies. Compared to conventional single-task fine-tuning and joint fine-tuning over all datasets without grouping, our RUS-based grouping strategy achieves consistently superior results across the majority of vision-language tasks. By organizing tasks according to their dominant multimodal interaction types—\textit{Redundancy}, \textit{Uniqueness}, and \textit{Synergy}—we enable the model to focus on shared reasoning patterns and avoid negative transfer from semantically disjoint tasks.

As shown in Fig.~\ref{fig:single}, \model\ also outperforms single-task fine-tuning, indicating that fine-tuning on a group of tasks sharing the same interaction type leads to more effective learning and improved performance. This suggests that aligned multimodal supervision within interaction-consistent task groups enhances representation learning and facilitates cross-task generalization.

Empirically, models fine-tuned with \model\ demonstrate stronger in-group generalization and better robustness compared to Unselective Multi-Task Fine-Tuning, especially on tasks requiring complex multimodal understanding, as shown in Fig.{\ref{fig:radar}}. These improvements confirm our hypothesis that interaction-type alignment provides a more meaningful signal than dataset labels alone. Notably, on tasks such as medical visual question answering and multimodal sentiment understanding, our grouped fine-tuning strategy yields substantial gains in both accuracy and generalization.

\begin{figure}[t]
  \centering
  \includegraphics[width=\textwidth]{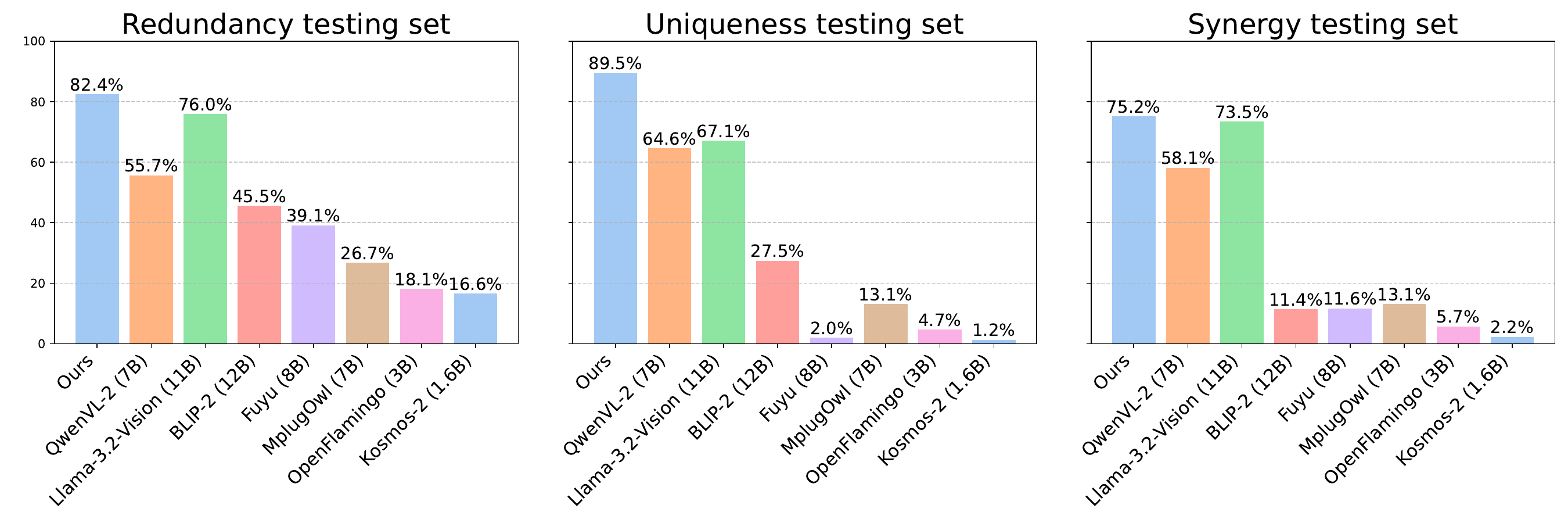} 
  \caption{\textbf{SoTA Open-source Models Comparison.} Our method outperforms other open-source models, including the base model Qwen2-VL (7B) and the recent LLaMA-3.2-Vision (11B). Our results are obtained by fine-tuning Qwen2-VL on grouped datasets corresponding to different interaction types.}
  \label{fig:opensource}
\end{figure}

\subsubsection{Comparisons to SoTA vision-language models}

From Fig.~\ref{fig:opensource}, our method surpasses the strongest open-source vision-language foundation models, including \kosmostwo~\citep{peng2023kosmos}, \openflamingo~\citep{awadalla2023openflamingo},  \mplugowl~\citep{ye2023mplug}, \fuyu~\citep{fuyu-8b}, \bliptwo~\citep{li2022blip}, and recent Llama-3.2-Vision-11B~\citep{grattafiori2024llama}. These models, while competitive in zero-shot and few-shot settings, do not benefit from targeted task grouping during fine-tuning. Current models can gain good performance on redundancy datasets, but they often perform badly on uniqueness and synergy datasets. Furthermore, when evaluated on a wide range of tasks involving image-text matching, visual question answering, diagram understanding, and affective computing, our RUS-guided fine-tuned model demonstrates substantial improvements in both accuracy and reliability. These gains are particularly prominent on complex tasks requiring fine-grained multimodal reasoning, such as scientific QA and visual sarcasm detection. Notably, the Llama-3.2-Vision-11B outperforms Qwen2-VL, but after task grouping and fine-tuning, \model\ outperforms both models.

These results emphasize that careful organization of the fine-tuning curriculum by grouping tasks based on their underlying interaction patterns can unlock capabilities beyond the largest pre-trained models. Our approach is complementary to large-scale pretraining, offering a principled fine-tuning strategy that can further enhance model performance.

\begin{table}[t]
\centering
\fontsize{8}{9}\selectfont
\caption{We report the performance of different task grouping strategies on each test dataset. These include our proposed \model\, as well as Unselective Multi-Task Fine-Tuning, where all 18 datasets are jointly used to fine-tune the base model (denoted as All). We also include results from fine-tuning with MixLoRA \citep{shen2024multimodal}, and from a similarity-based grouping strategy INSTA \citep{lee2024insta} that partitions the datasets into three groups based on task similarity.}
\label{tab:redundancy}
\begin{tabular}{lcccccc}
\toprule
Dataset & \makecell{\model} & \makecell{Unselective \\ Fine-Tuning}  &\makecell{MoE-based Group \\ Fine-tuning} & \multicolumn{3}{c}{\makecell{Similar Task Group \\ INSTA \citep{lee2024insta} Fine-tuning}}\\
\cmidrule(lr){2-2} \cmidrule(lr){3-3} \cmidrule(lr){4-4} \cmidrule(lr){5-7}
& \makecell{\textbf{Ours}} & All & MixLoRA \citep{shen2024multimodal} & Group 1 & Group 2 & Group 3 \\
\midrule
\textsc{nlvr} & \bf89.1 & 67.3 & 48.5 & 66.3 & 78.2 & 56.4 \\
\textsc{pathvqa} & \bf79.2 & 27.7 & 72.3 & 47.5 & 71.3 & 44.6 \\
\textsc{slake} & 71.3 & 35.6 & 44.6 & 57.4 & \bf74.3 & 73.2 \\
\textsc{vqa} & \bf90.1 & 47.5 & 57.4 & 45.5 & \bf90.1 & 78.2 \\
\midrule
\textsc{hatefulmemes} & \bf91.9 & 27.7 & 39.6 & 83.2 & 79.2 & 77.2 \\
\textsc{magicbrush} & \bf74.3 & 30.7 & 69.1 & 29.7 & 59.4 & 38.6 \\
\textsc{nycartoon} & \bf70.3 & 12.9 & 66.1 & 32.7 & 13.9 & 19.8 \\
\textsc{scienceqa} & \bf64.3 & 62.4 & 26.7 & 30.7 & 37.3 & 32.7 \\
\midrule
\textsc{inaturalist} & 86.4 & 0.0 & 80.1 & 80.2 & \bf93.1 & 40.6 \\
\textsc{lncoco} & \bf82.4 & 8.9 & 4.0 & 37.6 & 17.8 & 7.9 \\
\textsc{screen2words} & \bf89.1 & 23.8 & 78.2 & 70.3 & 44.6 & 85.1 \\
\textsc{ucmerced} & \bf100.0 & 46.5 &\bf 100.0 & 63.4 & \bf100.0 & 94.1 \\
\bottomrule
\end{tabular}
\end{table}

\subsubsection{Comparisons to alternative task groupings}

As shown in Tab.~\ref{tab:redundancy}, while both alternatives offer benefits over naive multitask tuning, they fall short of the performance achieved by our method. Similarity-based clustering tends to rely on surface-level linguistic similarity, which may conflate tasks with different interaction demands (e.g., visual-only vs. joint visual-linguistic reasoning). MixLora, on the other hand, introduces significant architectural complexity and overhead, and its routing decisions may not align well with the semantic structures critical for multimodal alignment. In contrast, \model\ yields better performance while maintaining a straightforward tuning pipeline, without the need for additional expert routing or gating mechanisms. Our results indicate that the simplicity and principled foundation of interaction-based grouping make it more effective and interpretable than existing alternatives.

\section{Conclusion}
\label{sec:conclusion}
This work introduced a novel task-grouping strategy for multimodal instruction tuning, based explicitly on multimodal interaction type—redundancy, uniqueness, and synergy. We leverage these intrinsic multimodal interactions to group tasks effectively, facilitating stronger intra-group transfer and reducing cross-task interference. Extensive experiments on the large-scale HEMM benchmark demonstrate that our RUS-based grouping consistently surpasses conventional single-task fine-tuning and indiscriminate multi-task fine-tuning methods. Additionally, it significantly outperforms alternative task grouping strategies such as instruction-text similarity clustering and MOE-based models. Our findings indicate that aligning fine-tuning tasks based on multimodal interaction types not only improves model specialization and robustness but also enhances generalization to unseen tasks. This principled approach offers a practical and interpretable method to optimize multimodal foundation models, highlighting its potential for broader application and further research.

{\footnotesize
\bibliographystyle{plainnat}
\bibliography{refs}
}

\clearpage
\appendix
\section{\model\  details}
We fine-tune our \model\ based on the powerful Qwen2-VL with LLaMA-Factory. In each interaction group, we follow the original setting, which utilize hyperparameters as Table. \ref{tab:hyperparams} shows, and it takes nearly one day to train on an NVIDIA A100 GPU.
\begin{table}[h]
  \centering
  \begin{tabular}{lcc}
    \toprule
    \textbf{Parameter} & \textbf{Value} \\
    \midrule
    Finetuning type        & LoRA \\
    Per device train batch size         & 2      \\
    Learning rate              & $1\times10^{-4}$ \\
    LR scheduler type        & cosine     \\
    Number of train epochs           & 3     \\
    Warmup Ratio                 & 0.1 \\
    Val size                     & 0.1 \\
    \bottomrule
  \end{tabular}
  \caption{Training hyperparameters.}
  \label{tab:hyperparams}
\end{table}
\section{Additional experiment}
\begin{figure}[htbp]
  \centering
  \includegraphics[width=\linewidth]{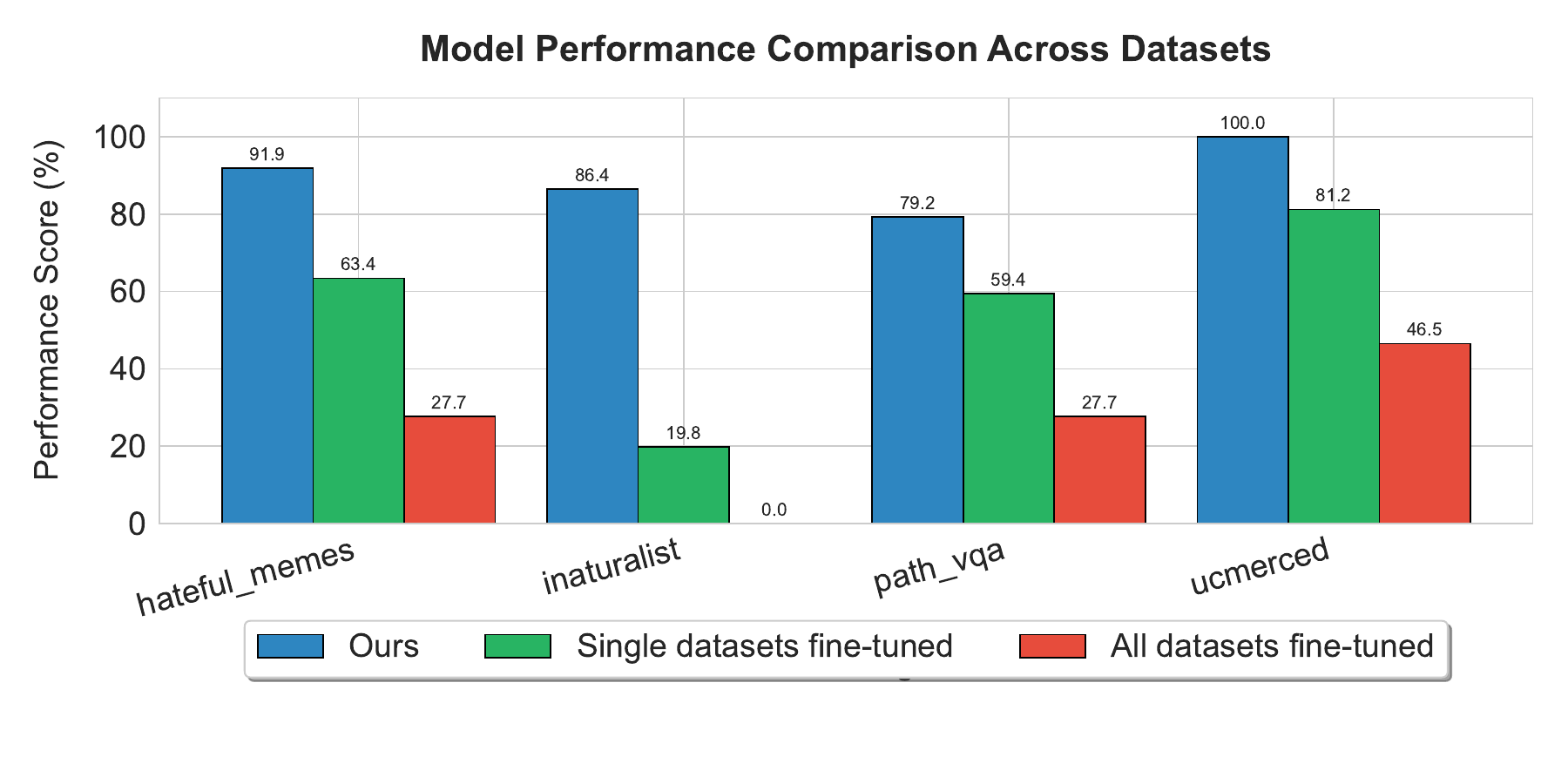}
  \caption{%
    \textbf{Model performance comparison across datasets.}
    ``Single datasets fine-tuned’’ uses only in-domain data that exactly matches each test set, while ``All datasets fine-tuned’’ mixes every dataset together without filtering.
    Our \model\ method selectively aggregates related datasets and consistently yields the highest accuracy, indicating that the proposed domain classification introduces beneficial cross-domain information.
    Conversely, indiscriminate aggregation dilutes task-relevant signals and harms performance, underscoring the necessity of careful dataset grouping.
  }
  \label{fig:performance_comparison}
\end{figure}

Fig.~\ref{fig:performance_comparison} contrasts three adaptation strategies across four representative benchmarks.  
\emph{Single-task fine-tuning} (green) uses training data drawn exclusively from the same domain as the test set; this establishes a strong in-domain baseline.  
Our \model\ (blue) additionally incorporates examples from other semantically related datasets, yet still surpasses the single-task baseline on every benchmark—demonstrating that our grouping introduces complementary information rather than harmful noise.  
In stark contrast, naively fine-tuning on \emph{all} available datasets without any selection (red) degrades performance, confirming that indiscriminate data mixing can obscure task-specific signals.  
These results validate the effectiveness of our domain classification strategy: by admitting only \emph{informative} auxiliary data, the model learns more transferable representations and achieves superior generalization.

\section{Individual dataset details}
\label{app:dataset_details}
In this section, we provide the details of the tasks and datasets chosen for the HEMM benchmark: we describe the split used to evaluate the models, any preprocessing applied to the samples, and their access restrictions and licenses.
\begin{enumerate}
   \item \textbf{\VQA} is a benchmark dataset comprising pairs of images and corresponding free-form, open-ended questions. Answering these questions often requires fine-grained recognition of objects and activities within the image, and in some cases, commonsense reasoning. A significant portion of the questions are binary in nature, typically requiring "yes" or "no" answers.

\textbf{Split:} We conduct evaluation on the validation split of real images, which contains a total of 244,302 questions.

\textbf{Prompt used:} You are given an image and a question. Answer the question in a single word. Question: \textit{<question>}

\textbf{Access restrictions:} The dataset can be downloaded from \href{https://visualqa.org/vqa_v1_download.html}{https://visualqa.org/vqa\_v1\_download.html}.

\textbf{Licenses:} All images in the dataset are provided under the CC BY 4.0 DEED license \href{https://creativecommons.org/licenses/by/4.0/deed.en}{https://creativecommons.org/licenses/by/4.0/deed.en}.

\textbf{Ethical considerations:} The dataset does not contain any personally identifiable information or offensive content.

    \item \textbf{\Decimer} dataset is a hand-drawn molecule image dataset consisting of chemical structure as the images and their SMILES representation as the strings. This SMILES representation stands for 'Simplified Molecular Input Line Entry System', which depicts the three-dimensional structure of the chemical into a string of symbols. In order to solve this task, the model should have an understanding of structure of the chemical and how these structures are depicted in the given format. 

    \textbf{Split:} The dataset consists of 5088 images over which evaluation has been performed.
    
    \textbf{Prompt used:} Simplified molecular-input line-entry system (SMILES) notation of the given molecule: 
    
    \textbf{Access restrictions:} The dataset is available to download from \href{https://zenodo.org/records/7617107}{https://zenodo.org/records/7617107}
    
    \textbf{Licenses:} The dataset is available under Creative Commons Attribution 4.0 International License \href{https://creativecommons.org/licenses/by/4.0/deed.en}{https://creativecommons.org/licenses/by/4.0/deed.en}, which permits use and sharing of data.

    \textbf{Ethical considerations:} No personally identifiable information or offensive content present in the dataset.

    \item \textbf{\Decimer} is a dataset of hand-drawn molecular structure images, each paired with its corresponding SMILES (Simplified Molecular Input Line Entry System) representation. The SMILES format encodes the three-dimensional molecular structure into a linear string of symbols. Solving this task requires the model to understand both the visual depiction of molecular structures and their translation into the SMILES notation.

\textbf{Split:} Evaluation is conducted on a set of 5,088 images.

\textbf{Prompt used:} Simplified molecular-input line-entry system (SMILES) notation of the given molecule:

\textbf{Access restrictions:} The dataset is publicly available at \href{https://zenodo.org/records/7617107}{https://zenodo.org/records/7617107}.

\textbf{Licenses:} This dataset is released under the Creative Commons Attribution 4.0 International License \href{https://creativecommons.org/licenses/by/4.0/deed.en}{https://creativecommons.org/licenses/by/4.0/deed.en}, allowing for broad use and redistribution.

\textbf{Ethical considerations:} The dataset does not contain any personally identifiable information or offensive content.

    \item \textbf{\ScienceQA} is a multiple-choice question dataset covering diverse science domains, including natural science, social science, and language science. Each question requires the model to select the correct answer from a given set of options. Supplementary materials such as lecture notes and explanations are optionally provided to aid in reasoning. While some questions lack visual content, we restrict our evaluation to the subset of questions that include an associated image.

\textbf{Split:} Evaluation is conducted on 4.24k questions from the test set.

\textbf{Prompt used:} You are given a question and a set of answer choices. Contextual information and an image are provided to assist in understanding the question. Additionally, lecture notes may be available to support your reasoning. Your task is to choose the best answer from the comma-separated choices. Return the selected choice exactly as it appears.  
\textit{lecture: <lecture>}  
\textit{question: <question>}  
\textit{context: <context>}  
\textit{choices: <choices>}  
Answer:

\textbf{Access restrictions:} The dataset is publicly available at \href{https://huggingface.co/datasets/derek-thomas/ScienceQA}{https://huggingface.co/datasets/derek-thomas/ScienceQA}.

\textbf{Licenses:} The dataset is released under the CC BY-NC-SA 4.0 license \href{https://creativecommons.org/licenses/by-nc-sa/4.0/deed.en}{https://creativecommons.org/licenses/by-nc-sa/4.0/deed.en}, which permits sharing under attribution, non-commercial use, and share-alike terms.

\textbf{Ethical considerations:} The dataset contains no personally identifiable information or offensive content.

    \item \textbf{\Slake} is a medical visual question-answering dataset that consists of image and question-answer pairs. Annotations have been done by experienced physicians and a medical knowledge base for medical visual question answering. The dataset consists of Yes/No type of questions as well as questions which could be answered with a single word. 

    \textbf{Split:} We use the test set of this dataset which consists of 2070 questions.

    \textbf{Prompt used:} Answer the question in a single word, Question: \textit{<question>}

     \textbf{Access restrictions:} The dataset is available to download from \href{https://huggingface.co/datasets/BoKelvin/SLAKE}{https://huggingface.co/datasets/BoKelvin/SLAKE}

    \textbf{Licenses:} Images under this dataset are available in CC-BY-SA 4.0 license \href{https://creativecommons.org/licenses/by-sa/4.0/deed.en}{https://creativecommons.org/licenses/by-sa/4.0/deed.en} which allows sharing data.

    \textbf{Ethical considerations:} No personally identifiable information or offensive content is present in the dataset.

    \item \textbf{\Slake} is a medical visual question answering (VQA) dataset comprising image-question-answer triplets. Annotations were provided by experienced physicians and curated using a medical knowledge base, ensuring domain-specific accuracy. The questions include both binary (Yes/No) types and those that can be answered with a single word.

\textbf{Split:} We evaluate on the test split, which contains 2,070 questions.

\textbf{Prompt used:} Answer the question in a single word.  
Question: \textit{<question>}

\textbf{Access restrictions:} The dataset is publicly available at \href{https://huggingface.co/datasets/BoKelvin/SLAKE}{https://huggingface.co/datasets/BoKelvin/SLAKE}.

\textbf{Licenses:} The dataset is released under the CC BY-SA 4.0 license \href{https://creativecommons.org/licenses/by-sa/4.0/deed.en}{https://creativecommons.org/licenses/by-sa/4.0/deed.en}, which permits data sharing under attribution and share-alike terms.

\textbf{Ethical considerations:} The dataset contains no personally identifiable information or offensive content.

    \item \textbf{\UCMercedLandUse} is a dataset for land use classification, comprising aerial images categorized into 21 distinct classes. The images were manually extracted from the USGS National Map Urban Area Imagery, covering various urban regions across the United States. All possible class labels are included in the prompt to allow the model to choose the appropriate category.

\textbf{Split:} Evaluation is performed on the validation split from \href{https://www.kaggle.com/datasets/apollo2506/landuse-scene-classification}{https://www.kaggle.com/datasets/apollo2506/landuse-scene-classification}.

\textbf{Prompt used:} You are given an image. Classify whether it belongs to one of the following categories: mediumresidential, buildings, tenniscourt, denseresidential, baseballdiamond, intersection, harbor, parkinglot, river, overpass, mobilehomepark, runway, forest, beach, freeway, airplane, storagetanks, chaparral, golfcourse, sparseresidential, agricultural. Choose one class from the list.

\textbf{Access restrictions:} The dataset can be downloaded from either \href{http://weegee.vision.ucmerced.edu/datasets/landuse.html}{http://weegee.vision.ucmerced.edu/datasets/landuse.html} or \href{https://www.kaggle.com/datasets/apollo2506/landuse-scene-classification}{https://www.kaggle.com/datasets/apollo2506/landuse-scene-classification}.

\textbf{Licenses:} No explicit license is provided for this dataset.

\textbf{Ethical considerations:} The dataset contains no personally identifiable information or offensive content.

    \item \textbf{\Enrico} is a topic classification dataset for mobile user interface (UI) screenshots. It builds upon the RICO dataset~\cite{deka2017rico}, with additional human annotations rating each UI design as either good or bad. Each sample is associated with a UI class—such as calculator, camera, chat, news, or profile—from which the model must select the most appropriate category based on the given image.

\textbf{Split:} Evaluation is conducted on the dataset available at \href{http://userinterfaces.aalto.fi/enrico/resources/screenshots.zip}{http://userinterfaces.aalto.fi/enrico/resources/screenshots.zip}.

\textbf{Prompt used:} You are given a screenshot of a mobile application's user interface. Choose the most appropriate design topic from the following comma-separated options: bare, dialer, camera, chat, editor, form, gallery, list, login, maps, mediaplayer, menu, modal, news, other, profile, search, settings, terms, tutorial.

\textbf{Access restrictions:} The dataset is publicly available at \href{https://github.com/luileito/enrico}{https://github.com/luileito/enrico}.

\textbf{Licenses:} The dataset is released under the MIT license \href{https://github.com/luileito/enrico/blob/master/LICENSE}{https://github.com/luileito/enrico/blob/master/LICENSE}.

\textbf{Ethical considerations:} The dataset contains no personally identifiable information or offensive content.

    \item \textbf{\MMIMDb} is a genre classification dataset consisting of movie posters and corresponding plot descriptions. Each movie can belong to multiple genres. The dataset was constructed using the MovieLens 20M dataset~\cite{harper2015movielens}, from which metadata such as genre, plot, release year, and other information were aggregated. For our evaluation, we use only the poster image and the plot to predict the associated genres.

\textbf{Split:} Evaluation is conducted on the test split.

\textbf{Prompt used:} You are given a movie poster and its corresponding plot. Select the appropriate genres from the following comma-separated list: drama, comedy, romance, thriller, crime, action, adventure, horror, documentary, mystery, sci-fi, fantasy, family, biography, war, history, music, animation, musical, western, sport, short, film-noir.  
Plot: \textit{<plot>}  
Note: A movie may belong to multiple genres. Provide all applicable genres, separated by commas.

\textbf{Access restrictions:} The dataset is publicly available for research use at \href{http://lisi1.unal.edu.co/mmimdb/}{http://lisi1.unal.edu.co/mmimdb/} and \href{https://github.com/johnarevalo/gmu-mmimdb/}{https://github.com/johnarevalo/gmu-mmimdb/}.

\textbf{Licenses:} The dataset is released under the MIT license \href{https://github.com/johnarevalo/gmu-mmimdb/blob/master/LICENSE}{https://github.com/johnarevalo/gmu-mmimdb/blob/master/LICENSE}.

\textbf{Ethical considerations:} The dataset contains no personally identifiable information or offensive content.

    \item \textbf{\VQARAD} is a visual question answering dataset based on radiology images. The images are sourced from MedPix\footnote{\url{https://medpix.nlm.nih.gov/home}}, an open-access radiology image database. The dataset was manually curated by clinical annotators, including medical students and senior radiologists. Ground truth answers cover a variety of question types, such as those related to counting, color, abnormalities, and the presence of specific conditions.

\textbf{Split:} Evaluation is conducted on the test set available at \href{https://huggingface.co/datasets/flaviagiammarino/vqa-rad/viewer/default/test}{https://huggingface.co/datasets/flaviagiammarino/vqa-rad/viewer/default/test}, which includes 451 questions.

\textbf{Prompt used:} You are given a radiology image and a question. Answer the question in a single word.  
Question: \textit{<question>}

\textbf{Access restrictions:} The dataset is publicly available at \href{https://huggingface.co/datasets/flaviagiammarino/vqa-rad/viewer}{https://huggingface.co/datasets/flaviagiammarino/vqa-rad/viewer}.

\textbf{Licenses:} The dataset is released under the Creative Commons Attribution 4.0 International License \href{https://creativecommons.org/licenses/by/4.0/deed.en}{https://creativecommons.org/licenses/by/4.0/deed.en}.

\textbf{Ethical considerations:} The dataset contains no personally identifiable information or offensive content.

    \item \textbf{\FlickrThirtyK} is an image captioning dataset sourced from Flickr\footnote{\url{https://www.flickr.com/}}, extending the dataset introduced by~\cite{hodosh2013framing}. It follows similar data collection and annotation protocols, providing a rich set of images with corresponding human-written captions.

\textbf{Split:} Evaluation is conducted on the test split.

\textbf{Prompt used:} A picture of \textit{<image description>}.

\textbf{Access restrictions:} The dataset is publicly available at \href{https://www.kaggle.com/datasets/hsankesara/flickr-image-dataset}{https://www.kaggle.com/datasets/hsankesara/flickr-image-dataset}.

\textbf{Licenses:} The dataset is released under the CC0 Public Domain license \href{https://creativecommons.org/publicdomain/zero/1.0/deed.en}{https://creativecommons.org/publicdomain/zero/1.0/deed.en}.

\textbf{Ethical considerations:} The dataset contains no personally identifiable information or offensive content.

    \item \textbf{\FER} is a widely used dataset for facial expression recognition, where each facial image is classified into one of seven emotion categories. The images were collected using the Google Search API, and OpenCV was employed to detect and extract face bounding boxes.

\textbf{Split:} Evaluation is conducted on the test split available at \href{https://www.kaggle.com/datasets/msambare/fer2013}{https://www.kaggle.com/datasets/msambare/fer2013}.

\textbf{Prompt used:} Given a photo of a face, determine the facial expression. Choose from the following options: angry, disgust, fear, happy, neutral, sad, surprise. Answer in a single word.

\textbf{Access restrictions:} The dataset can be downloaded from \href{https://www.kaggle.com/datasets/msambare/fer2013}{https://www.kaggle.com/datasets/msambare/fer2013}.

\textbf{Licenses:} No license is explicitly provided with the dataset.

\textbf{Ethical considerations:} The dataset contains facial images collected via Google Image Search, but no personally identifiable information or offensive content is included.

    \item \textbf{\NewYorkerCartoon} is collected from the weekly New Yorker magazine cartoon captioning contest \footnote{https://www.newyorker.com/cartoons/contest}, where readers are tasked to give a humorous caption for a cartoon image and the funniest captions are selected based on public votes. The dataset is formulated based on taking in the image and caption to predict how funny the pair is based on the normalized number of votes. Given an image and its caption, we ask the model if the caption is humorous or not. Each image has multiple caption choices with votes for the caption being not funny, somewhat funny, funny. We select the funniest caption to have a ground truth answer as 'yes' when prompted for evaluation. The next four funniest captions are selected to have ground truth answers as 'no' when prompted for evaluation. 
    
    \textbf{Split:} We use the data available on \href{https://github.com/nextml/caption-contest-data}{https://github.com/nextml/caption-contest-data}
    
    \textbf{Prompt used:} You are given a cartoon image and a caption. start the answer with yes if the caption is funny or No if the caption is not funny. Caption: \textit{<caption>}

    \textbf{Access restrictions:} The dataset is available to download from \href{https://github.com/nextml/caption-contest-data}{https://github.com/nextml/caption-contest-data}
    
    \textbf{Licenses:} No license is provided with the dataset.

    \textbf{Ethical considerations:} No personally identifiable information or offensive content is present in the dataset.
    
    \item \textbf{\NewYorkerCartoon} is a dataset derived from the weekly cartoon captioning contest hosted by *The New Yorker*\footnote{\url{https://www.newyorker.com/cartoons/contest}}, where readers submit humorous captions for cartoon images, and the most amusing entries are selected based on public votes. Each data sample pairs a cartoon with a caption, along with vote-based humor ratings categorized as “not funny,” “somewhat funny,” or “funny.” For evaluation, we formulate a binary classification task: given a cartoon and a caption, the model must determine whether the caption is funny. The caption with the highest number of votes is labeled as "yes," while the next four ranked captions are labeled as "no."

\textbf{Split:} We use the data available at \href{https://github.com/nextml/caption-contest-data}{https://github.com/nextml/caption-contest-data}.

\textbf{Prompt used:} You are given a cartoon image and a caption. Start your answer with “Yes” if the caption is funny, or “No” if it is not.  
Caption: \textit{<caption>}

\textbf{Access restrictions:} The dataset is publicly available at \href{https://github.com/nextml/caption-contest-data}{https://github.com/nextml/caption-contest-data}.

\textbf{Licenses:} No explicit license is provided with the dataset.

\textbf{Ethical considerations:} The dataset contains no personally identifiable information or offensive content.
    
    \item \textbf{\MagicBrush} is an instruction-based image editing dataset featuring both single-turn and multi-turn editing tasks. Each sample includes an image and corresponding textual instructions, enabling guided image manipulation. The images are sampled from the MS COCO dataset~\cite{lin2014microsoft}, and the edits were collected via crowdworkers on Amazon Mechanical Turk (AMT)\footnote{\url{https://www.mturk.com/}}, using DALL-E 2\footnote{\url{https://openai.com/dall-e-2}} for generation. For our evaluation, we focus exclusively on the single-turn instruction-editing subset.

\textbf{Split:} Evaluation is conducted on the test split available at \href{https://osu-nlp-group.github.io/MagicBrush/}{https://osu-nlp-group.github.io/MagicBrush/}.

\textbf{Prompt used:} Edit the given image based on the provided instruction.  
Instruction: \textit{<instruction>}

\textbf{Access restrictions:} The dataset is publicly available at \href{https://osu-nlp-group.github.io/MagicBrush/}{https://osu-nlp-group.github.io/MagicBrush/}.

\textbf{Licenses:} The dataset is released under the CC BY 4.0 license \href{https://creativecommons.org/licenses/by/4.0/deed.en}{https://creativecommons.org/licenses/by/4.0/deed.en}.

\textbf{Ethical considerations:} The dataset contains no personally identifiable information or offensive content.

    \item \textbf{\MemeCap} is a meme captioning dataset, with images sourced from the r/memes subreddit\footnote{\url{https://www.reddit.com/r/memes/}}. Captions were generated through a two-round annotation process by human workers on Amazon Mechanical Turk. For evaluation, the model is provided with the title and a description of the meme image, and is asked to infer the intended message or humor conveyed by the meme.

\textbf{Split:} Evaluation is conducted on the test set available at \href{https://github.com/eujhwang/meme-cap/tree/main}{https://github.com/eujhwang/meme-cap/tree/main}.

\textbf{Prompt used:} This is a meme with the title \textit{<title>}. The image description is \textit{<image\_description>}. What is the meme poster trying to convey?  
Answer:

\textbf{Access restrictions:} The dataset is publicly available at \href{https://github.com/eujhwang/meme-cap/tree/main}{https://github.com/eujhwang/meme-cap/tree/main}.

\textbf{Licenses:} No license is provided for the dataset.

\textbf{Ethical considerations:} While the dataset does not contain personally identifiable information, it may include offensive content due to the nature of meme data sourced from public internet platforms.

    \item \textbf{\HatefulMemes} is a multimodal classification dataset released as part of a challenge hosted by Meta, designed to assess whether a meme image paired with its textual caption expresses hateful intent. The images were sourced from Getty Images\footnote{\url{https://www.gettyimages.in/}} and annotated by a third-party platform. Each sample presents an image and an overlaid text phrase, which may only convey hateful meaning when interpreted together.

\textbf{Split:} Evaluation is conducted on the “dev” split, available at \href{https://www.kaggle.com/datasets/parthplc/facebook-hateful-meme-dataset/data}{https://www.kaggle.com/datasets/parthplc/facebook-hateful-meme-dataset/data}.

\textbf{Prompt used:} You are given an image. The image and the accompanying text phrase may appear innocuous individually, but together may convey a hateful message.  
Text phrase: \textit{<text\_phrase>}  
Judge whether the combination of image and text is hateful. Begin your answer with either "yes" or "no", where "yes" indicates the meme is hateful and "no" indicates it is not.  
Answer:

\textbf{Access restrictions:} The dataset is publicly available at \href{https://www.kaggle.com/datasets/parthplc/facebook-hateful-meme-dataset/data}{https://www.kaggle.com/datasets/parthplc/facebook-hateful-meme-dataset/data}.

\textbf{Licenses:} The images are covered under the Getty Images license \href{https://www.gettyimages.in/eula}{https://www.gettyimages.in/eula}.

\textbf{Ethical considerations:} While the dataset contains no personally identifiable information, it may include offensive content. This is intentional, as the dataset aims to enable models to detect and mitigate harmful multimodal content.

    \item \textbf{\iNaturalist} is a large-scale image classification dataset encompassing over 5,000 wildlife species of plants and animals. The images and labels are sourced from the iNaturalist platform\footnote{\url{https://www.inaturalist.org/}}. The task involves identifying the species depicted in a given image. Unlike other classification datasets, the full list of possible classes is not provided to the model due to the dataset’s broad taxonomic coverage.

\textbf{Split:} Evaluation is conducted on the validation split from the 2021 edition of the dataset.

\textbf{Prompt used:} The scientific species name of the species present in the image is:

\textbf{Access restrictions:} The dataset is available at \href{https://ml-inat-competition-datasets.s3.amazonaws.com/2021/val.tar.gz}{https://ml-inat-competition-datasets.s3.amazonaws.com/2021/val.tar.gz}.

\textbf{Licenses:} The dataset is distributed under the MIT license \href{https://github.com/visipedia/inat_comp/blob/master/LICENSE}{https://github.com/visipedia/inat\_comp/blob/master/LICENSE}.

\textbf{Ethical considerations:} The dataset contains no personally identifiable information or offensive content.

    \item \textbf{\NLVR} (Natural Language for Visual Reasoning) is a visual reasoning dataset consisting of image-text pairs. Each image is synthetically generated by randomly sampling objects and their properties. Crowdworkers are then asked to write natural language sentences describing these images, enabling fine-grained reasoning tasks.

\textbf{Split:} Evaluation is conducted on the development split, available at \href{https://github.com/lil-lab/nlvr}{https://github.com/lil-lab/nlvr}.

\textbf{Prompt used:} Given an image and a related question, answer with a single word: either “true” or “false.”  
Question: \textit{<question>}

\textbf{Access restrictions:} The dataset can be downloaded from \href{https://github.com/lil-lab/nlvr}{https://github.com/lil-lab/nlvr}.

\textbf{Licenses:} No license is explicitly provided with the dataset.

\textbf{Ethical considerations:} The dataset contains no personally identifiable information or offensive content.

    \item \textbf{\Resisc} is a remote sensing image classification dataset containing aerial land use scenes categorized into 45 distinct classes. The images were collected from Google Earth by experts in remote sensing image interpretation. During evaluation, all class names are included in the prompt, and the model is asked to select the most appropriate class.

\textbf{Split:} We evaluate using the dataset available at \href{https://www.kaggle.com/datasets/happyyang/nwpu-data-set}{https://www.kaggle.com/datasets/happyyang/nwpu-data-set}.

\textbf{Prompt used:} You are given an image. Classify whether it belongs to one of the following categories:  
\texttt{'basketball\_court', 'overpass', 'ground\_track\_field', 'church', 'chaparral', 'forest', 'parking\_lot', 'golf\_course', 'baseball\_diamond', 'meadow', 'beach', 'sparse\_residential', 'desert', 'terrace', 'palace', 'bridge', 'commercial\_area', 'stadium', 'runway', 'lake', 'railway', 'tennis\_court', 'ship', 'intersection', 'river', 'freeway', 'airplane', 'industrial\_area', 'mountain', 'storage\_tank', 'cloud', 'roundabout', 'wetland', 'mobile\_home\_park', 'island', 'harbor', 'railway\_station', 'medium\_residential', 'sea\_ice', 'thermal\_power\_station', 'snowberg', 'circular\_farmland', 'airport', 'dense\_residential', 'rectangular\_farmland'}.  
Choose a class from the above list.

\textbf{Access restrictions:} The dataset can be downloaded from \href{https://www.kaggle.com/datasets/happyyang/nwpu-data-set}{https://www.kaggle.com/datasets/happyyang/nwpu-data-set}.

\textbf{Licenses:} No license is explicitly provided for this dataset.

\textbf{Ethical considerations:} The dataset contains no personally identifiable information or offensive content.

    \item \textbf{Localized Narratives (COCO subset)} (\LNCOCO) is derived from the Localized Narratives dataset~\cite{pont2020connecting}, which augments images from COCO~\cite{lin2014microsoft}, Flickr30k~\cite{flickr30k}, and ADE20K~\cite{zhou2019semantic} with detailed spatial annotations and spoken captions. We use the COCO subset of this dataset for the task of image generation, where the model generates an image based on a textual description.

\textbf{Split:} We evaluate on the COCO subset, which contains 8,573 samples. Ground truth images are taken from the MS COCO 2017 validation set.

\textbf{Prompt used:} Generate an image based on the provided caption.  
Caption: \textit{<caption>}

\textbf{Access restrictions:} The dataset is available at \href{https://google.github.io/localized-narratives/}{https://google.github.io/localized-narratives/}.

\textbf{Licenses:} The dataset is released under the CC BY-NC 4.0 license \href{https://creativecommons.org/licenses/by-nc/4.0/}{https://creativecommons.org/licenses/by-nc/4.0/}.

\textbf{Ethical considerations:} The dataset contains no personally identifiable information or offensive content.

\end{enumerate}

\section{Model Details}
\label{app:model_details}
For the HEMM benchmark, we currently evaluate the following models. All the models except for Gemini and GPT-4V are open source and we encourage the community to add more models to the benchmark.

\begin{enumerate}
    \item \textbf{\bliptwo} is a vision-language model that combines a pre-trained image encoder and a pre-trained large language model (LLM) via a lightweight Q-former module. The Q-former uses an attention mechanism to fuse image queries with input text, producing a joint representation that is passed to the decoder for response generation. During supervised fine-tuning, only the Q-former parameters are updated, while the vision encoder and language decoder remain frozen.

In our experiments, we use the \verb|blip2_t5| model with the \verb|pretrain_flant5xxl| decoder, as implemented in the LAVIS framework\footnote{\url{https://github.com/salesforce/LAVIS/tree/main/projects/blip2}}. The selected model contains 108M trainable parameters and a total of 12.1B parameters.

\textbf{License:} The model is released under the BSD-3-Clause License \href{https://github.com/salesforce/LAVIS/blob/main/LICENSE.txt}{https://github.com/salesforce/LAVIS/blob/main/LICENSE.txt}.

\textbf{Access restrictions:} The model is publicly available through the LAVIS repository at \href{https://github.com/salesforce/LAVIS}{https://github.com/salesforce/LAVIS}.

    \item \textbf{\openflamingo} is an open-source implementation of the Flamingo model~\cite{alayrac2022flamingo}, designed for multimodal reasoning over sequences of interleaved images and texts. Unlike models limited to single-image inputs (e.g., BLIP2 or MiniGPT-4), \openflamingo supports multi-image processing within a single sample. The architecture integrates pre-trained vision and language components, where outputs from the vision encoder are injected into the language model layers via cross-modal attention. During training, all pre-trained components remain frozen except the cross-modal attention layers.

For evaluation, we use the \verb|OpenFlamingo-3B-vitl-mpt1b| model from the OpenFlamingo GitHub repository\footnote{\url{https://github.com/mlfoundations/open_flamingo}}, which includes 1.4B trainable parameters and 3.2B total parameters. The model is trained on 180 million image-text pairs.

\textbf{License:} The model is released under the MIT License \href{https://github.com/mlfoundations/open_flamingo/blob/main/LICENSE}{https://github.com/mlfoundations/open\_flamingo/blob/main/LICENSE}.

\textbf{Access restrictions:} The model is publicly available at \href{https://github.com/mlfoundations/open_flamingo}{https://github.com/mlfoundations/open\_flamingo}.


    \item \textbf{\fuyu} is a decoder-only transformer model that processes visual inputs by linearly projecting image patches into the first layer of the decoder. Its architecture is identical to that of Persimmon-8B\footnote{\url{https://www.adept.ai/blog/persimmon-8b}}, and thus inherits its model categorization. Persimmon-8B contains 9.3 billion parameters and was trained from scratch.

In this work, we evaluate the \verb|Fuyu-8B| model available on Hugging Face\footnote{\url{https://huggingface.co/adept/fuyu-8b}}. As instruction-tuned versions of the model are not yet available, we rely on the base pre-trained version. The specific pre-training data sources and dataset sizes have not been disclosed.

\textbf{License:} The model is released under the Creative Commons Attribution-NonCommercial 4.0 International License \href{https://spdx.org/licenses/CC-BY-NC-4.0}{https://spdx.org/licenses/CC-BY-NC-4.0}.

\textbf{Access restrictions:} The model is publicly available at \href{https://huggingface.co/adept/fuyu-8b}{https://huggingface.co/adept/fuyu-8b}.

    \item \textbf{\kosmostwo} is a causal Transformer-based language model, extending the architecture of Kosmos-1~\cite{huang2024language}. It is trained using the next-token prediction objective. In addition to the original pre-training data used in Kosmos-1, \kosmostwo incorporates grounded image-text pairs for improved multimodal alignment. The model is first pre-trained on interleaved image-text data and subsequently instruction-tuned with both multimodal and language-only instructions.

For our experiments, we evaluate the \verb|ydshieh/kosmos-2-patch14-224| checkpoint available on Hugging Face\footnote{\url{https://huggingface.co/microsoft/kosmos-2-patch14-224}}, which contains 1.6 billion parameters.

\textbf{License:} The model is released under the MIT License \href{https://huggingface.co/datasets/choosealicense/licenses/blob/main/markdown/mit.md}{https://huggingface.co/datasets/choosealicense/licenses/blob/main/markdown/mit.md}.

\textbf{Access restrictions:} The model is publicly available at \href{https://huggingface.co/microsoft/kosmos-2-patch14-224}{https://huggingface.co/microsoft/kosmos-2-patch14-224}.

    \item \textbf{\mplugowl} is a multimodal model that integrates visual and textual inputs using a staged fusion strategy. A vision foundation model encodes the input image, and a visual abstractor module summarizes the visual features. These abstracted representations are then combined with text queries and passed to a pre-trained language model for response generation.

The training process consists of two stages. In the first stage, all components except the language model are fine-tuned using supervised learning. In the second stage, the language model is instruction-tuned on both multimodal and language-only instructions, while the rest of the model remains frozen.

We evaluate the model from the mPLUG-Owl GitHub repository\footnote{\url{https://github.com/X-PLUG/mPLUG-Owl/tree/main/mPLUG-Owl}}, which contains 7.2 billion parameters.

\textbf{License:} The model is released under the MIT License \href{https://github.com/X-PLUG/mPLUG-Owl/blob/main/LICENSE}{https://github.com/X-PLUG/mPLUG-Owl/blob/main/LICENSE}.

\textbf{Access restrictions:} The model is publicly available at \href{https://github.com/X-PLUG/mPLUG-Owl}{https://github.com/X-PLUG/mPLUG-Owl}.

    \item \textbf{Qwen2-VL} adopts a two-tower architecture, in which a ViT-based image encoder extracts visual features that are then aligned with the language space via a lightweight projection module. The resulting visual tokens are interleaved with textual inputs and passed to a pre-trained Qwen2 language model, which serves as the decoder backbone.

The model is trained in multiple stages: initial multimodal pretraining on aligned image-text pairs, followed by instruction tuning on curated multimodal prompts for downstream tasks.

We evaluate the 1.8B parameter version available at \href{https://huggingface.co/Qwen/Qwen-VL}{https://huggingface.co/Qwen/Qwen-VL}.

\textbf{License:} The model is released under the Apache 2.0 License \href{https://huggingface.co/Qwen/Qwen-VL/blob/main/LICENSE}{https://huggingface.co/Qwen/Qwen-VL/blob/main/LICENSE}.

\textbf{Access restrictions:} The model is publicly available via \href{https://huggingface.co/Qwen/Qwen-VL}{https://huggingface.co/Qwen/Qwen-VL}.

    \item \textbf{LLaMA3.2-V} extends the LLaMA 3.2 language model with vision capabilities via a gated multimodal adapter. A vision encoder—typically a CLIP ViT model—extracts dense image embeddings, which are projected and injected into the language model through cross-attention layers. Compared to earlier architectures, LLaMA3.2-V supports multi-image input, region-level grounding, and fine-grained visual reasoning.

The model is instruction-tuned using a diverse set of tasks that include visual, textual, and multimodal instructions. We evaluate the \verb|LLaMA-3.2-Vision-11B| checkpoint.

\textbf{License:} Released under the Meta Research License (non-commercial use) \href{https://ai.meta.com/resources/models-and-libraries/llama-3}{https://ai.meta.com/resources/models-and-libraries/llama-3}.

\textbf{Access restrictions:} Model weights are available upon request and subject to Meta’s usage terms, as listed at \href{https://huggingface.co/meta-llama/Llama-3.2-11B-Vision-Instruct}{https://huggingface.co/meta-llama/Llama-3.2-11B-Vision-Instruct}.

\end{enumerate}


\newpage
\appendix
\onecolumn

\end{document}